\documentclass[letterpaper]{article} % DO NOT CHANGE THIS
\usepackage{aaai23}  % DO NOT CHANGE THIS
\usepackage{times}  % DO NOT CHANGE THIS
\usepackage{helvet}  % DO NOT CHANGE THIS
\usepackage{courier}  % DO NOT CHANGE THIS
\usepackage[hyphens]{url}  % DO NOT CHANGE THIS
\usepackage{graphicx} % DO NOT CHANGE THIS
\usepackage{natbib}  % DO NOT CHANGE THIS AND DO NOT ADD ANY OPTIONS TO IT
\usepackage{caption} % DO NOT CHANGE THIS AND DO NOT ADD ANY OPTIONS TO IT
\usepackage{amssymb}
\usepackage{amsmath}
\usepackage{amsthm}
\usepackage{pifont}% http://ctan.org/pkg/pifont
\newcommand{\cmark}{\ding{51}}%
\newcommand{\xmark}{\ding{55}}%
\newtheorem{myDef}{Preliminary}
\newtheorem{myproblem}{Problem}
  
\usepackage{algorithm}
\usepackage{colortbl}
\usepackage{xcolor}
\usepackage{makecell}
\usepackage{booktabs}
\usepackage{multirow}
\usepackage{amssymb}
\usepackage{algpseudocode}  
\usepackage{amsmath} %
\usepackage{color}

\usepackage{newfloat}
\usepackage{listings}

\usepackage{arydshln}
\DeclareCaptionStyle{ruled}{labelfont=normalfont,labelsep=colon,strut=off} % DO NOT CHANGE THIS
\floatstyle{ruled}
\newfloat{listing}{tb}{lst}{}
\floatname{listing}{Listing}
\title{GANTEE: Generative Adversatial Network for Taxonomy Entering Evaluation}
\author{
    Zhouhong Gu\textsuperscript{\rm 1,4},
    Sihang Jiang\textsuperscript{\rm 1},
    Jingping Liu\textsuperscript{\rm 2}\thanks{Corresponding Authors},
    Yanghua Xiao\textsuperscript{\rm 1,3*},
    Hongwei Feng\textsuperscript{\rm 1},
    Zhixu Li\textsuperscript{\rm 1},\\
    Jiaqing Liang\textsuperscript{\rm 4},
    Jian Zhong\textsuperscript{\rm 5}
}
\affiliations{
    \textsuperscript{\rm 1}Shanghai Key Laboratory of Data Science, School of Computer Science, Fudan University, China\\
    \textsuperscript{\rm 2}School of Information Science and Engineering, East China University of Science and Technology\\
    \textsuperscript{\rm 3}Fudan-Aishu Cognitive Intelligence Joint Research Center\\
    \textsuperscript{\rm 4}School of Data Science, Fudan University\\
    \textsuperscript{\rm 5}HUAWEI CBG Edu AI Lab\\
    \{zhgu20, shawyh, hwfeng, zhixuli, liangjiaqing\}@fudan.edu.cn\\
    tedsihangjiang@gmail.com, 
    jingpingliu@ecust.edu.cn, 
    zhongjian6@huawei.com
}

\usepackage{bibentry}
\begin{document}

\maketitle
\vspace{-3mm}

\begin{abstract}
Taxonomy is formulated as directed acyclic concepts graphs or trees that support many downstream tasks.
Many new coming concepts need to be added to an existing taxonomy.
The traditional taxonomy expansion task aims only at finding the best position for new coming concepts in the existing taxonomy. 
However, they have two drawbacks when being applied to the real-scenarios.
The previous methods suffer from low-efficiency since they waste much time when most of the new coming concepts are indeed noisy concepts. They also suffer from low-effectiveness since they collect training samples only from the existing taxonomy, which limits the ability of the model to mine more hypernym-hyponym relationships among real concepts.
This paper proposes a pluggable framework called Generative Adversarial Network for Taxonomy Entering Evaluation (GANTEE) to alleviate these drawbacks.
A generative adversarial network is designed in this framework by discriminative models to alleviate the first drawback and the generative model to alleviate the second drawback.
Two discriminators are used in GANTEE to provide long-term and short-term rewards, respectively.
Moreover, to further improve the efficiency, pre-trained language models are used to retrieve the representation of the concepts quickly.
The experiments on three real-world large-scale datasets with two different languages show that GANTEE improves the performance of the existing taxonomy expansion methods in both effectiveness and efficiency.

\end{abstract}

\section{Introduction}
Taxonomy is formulated as directed acyclic graphs or trees, which consist of hyponym-hypernym relations between concepts.
One concept is a hypernym of another concept if the meaning of the former covers the latter~\cite{sang2007extracting}.
A well-constructed taxonomy assist various downstream tasks, including web content tagging~\cite{liu2020giant,liu2019tencent_taxonomy,peng2019RCNN4taxonomy}, personalized recommendation~\cite{karamanolakis2020txtract, huang2019taxonomy}, query understanding~\cite{yang2020co} and so on.

Manually maintaining a taxonomy is labor-intensive and time-consuming.
For example, millions of new concepts are expected to be added to the taxonomy of one of the largest shopping platforms in the world each month~\cite{jiang2019towards}.
So the task of automated taxonomy expansion is proposed~\cite{jurgens2016semeval}, which aims to automatically assign an existing concept (anchor concept) as a hypernym concept to the newly input concept (query concept).
These models successfully save much time from labor annotation in taxonomy maintenance.
These methods predict the hypernym-hyponym relationships mainly based on the representation similarity, so they sample the training data based on the existing taxonomy to learn good representations for each concept. 

However, we argue that these methods~\cite{yu2020steam,wang2022qen,cheng2022learning,shen2020taxoexpan,mao2020octet,wang2021enquire,zhang2021taxonomy} have two drawbacks when being applied to real applications. 
\textbf{First}, these methods suffer from low effectiveness in real applications due to limited ability to represent the semantics of concepts. Only sampling from existing taxonomy provides limited data to train representations for zero-shot concepts, where thousands times queries of the number of concepts in existing taxonomy are all unseen to the models~\cite{cheng2022learning}.
\textbf{Second}, these methods suffer from low efficiency in real applications because most of the query concepts are noisy.
For instance, in a leading ordering take-outs online platform, only thousands of concepts out of billions of queries are finally added to their taxonomy~\cite{cheng2022learning}.
Previous methods ignore the noisy concepts and waste much time finding the anchor concept for noisy query concepts.

\begin{figure*}
    \centering
    \resizebox{1\textwidth}{!}{
    \includegraphics{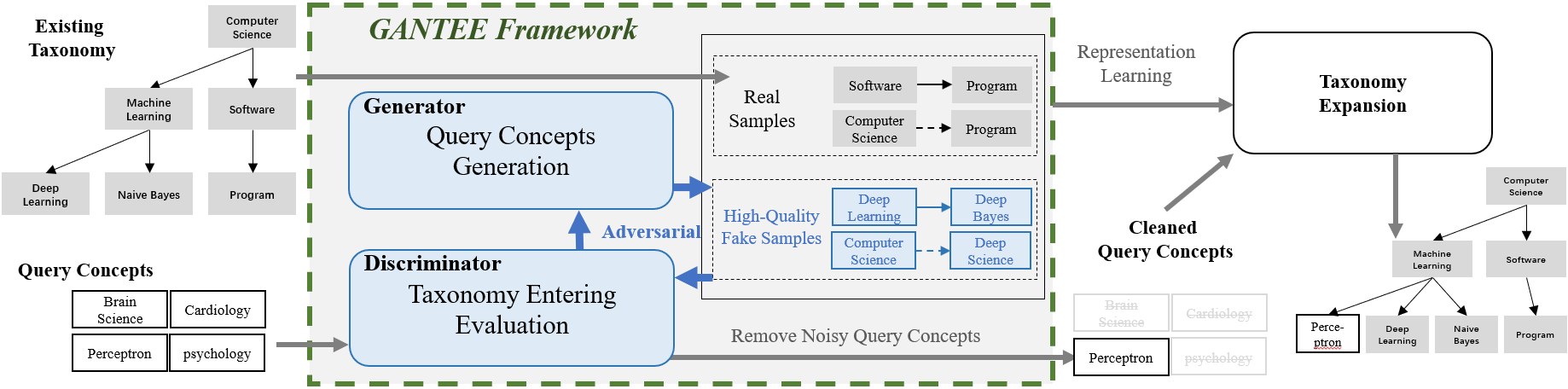}}
    \caption{An example of the GANTEE framework. This framework is implemented as a plugin before taxonomy expansion. The demand for generating high-quality data and taxonomy entering evaluation is formulated as a generative adversarial network in GANTEE.}
    \label{fig:introduction}
    \vspace{-7mm}
\end{figure*}
In this paper, we propose a pluggable framework called \textbf{G}enerative \textbf{A}dversarial \textbf{N}etwork for \textbf{T}axonomy \textbf{E}nterance \textbf{E}valuation (GANTEE) to boost both the effectiveness and efficiency of the taxonomy expansion methods in real-scenarios as is shown in Figure \ref{fig:introduction}.
To improve the effectiveness, GANTEE proposes a generative model to generate more high-quality training data, which are required to train the high-quality representation for taxonomy expansion. 
To improve efficiency, GANTEE introduces a new task called taxonomy entering evaluation, which removes many noisy query concepts before finding the suitable anchor concept for them.
Intuitively, the generative model and the taxonomy entering evaluation model are two adversarial models which can improve each other. The generative model produces fake samples similar to real ones to confuse the discriminative model. 
The taxonomy entering evaluation model learns to generate high-quality samples. Therefore, these two tasks can be formulated as a generative adversarial task~\cite{goodfellow2014generative} to improve their performance further.

Specifically, to improve efficiency, the taxonomy entering evaluation model is expected to be more lightweight than the taxonomy expansion model. 
We introduce two mechanisms to ensure the efficiency of the taxonomy entering evaluation model. 
First, instead of determining where to extend the query concept, GANTEE introduces two more manageable tasks \emph{``isA Concept Access''} and \emph{``isA Query Access''} to determine whether the query concept is a concept and whether the query concept should be added to the taxonomy.
Second, instead of training a new representation for emerging query concepts, we use pre-trained language models (PLMs) to directly obtain a representation of each concept based on the textual features of the concept. 
Although PLMs have limited ability in representing the fine-grained semantics based on conceptual text~\cite{lauscher2019specializing}, the experimental results verify that PLMs are suitable to generate representation in \emph{``isA Concept Access''} and \emph{``isA Query Access''} task.

In the experiments, we benchmark the taxonomy entering evaluation task on three real-world taxonomies in two languages, English and Chinese.
We reached the new SOTA in the experiments by improving all the metrics in only a third of the time of other methods in the prediction stage.
Finally, extensive experiments have been conducted to study how the parameters affect the result of GANTEE, and provide an effective tuning scheme for the following researchers who will use our method in the future.

\subsubsection{Contribution.}
To summarize, our major contributions include: 
(1) We propose a more manageable task called taxonomy entering evaluation which judges whether a query concept should be added to an existing taxonomy efficiently;
(2) We propose a novel, effective method called GANTEE to generate confusing negative and fidelity positive data boosts the performance and efficiency for both traditional taxonomy expansion tasks and taxonomy entering evaluation tasks.
(3) Extensive experiments have been conducted to verify the superiority of GANTEE on three datasets in two languages. We also propose an effective tuning method based on the experiment results.

% \begin{figure}
%     \centering
%     \includegraphics[width=1\linewidth]{Framework.png}
%     \caption{The illustration of GANTEE. 
%     The inputs are samples of the existing taxonomy and query concepts and outputs are high-quality generated samples and cleaned query concepts.
%     The generative model $G_\theta$ is used to generate fake query concepts based on given anchor concepts and labels.
%     Samples of the exsiting taxonomy are used to pre-train the generative model $G_\theta$.
%     Both generated and real samples are used to train the discriminative models $D_{H\phi}$ and $D_{R\phi}$.
%     If the samples are not finishing generating, then it is judged by the rollout discriminator.
%     Each sample is used to complete the possible generated sequences for the final reward using Monto Carlo Searching, and discriminative model $D_{R\phi}$ used to judge whether the generated text represent a concept.
%     If the samples generation are finished, then it is judged by the hyper discriminator, using the positional embeddings $pos_a$ and $pos_q$ to connect the representations of the anchor concept and query concept, and using the discriminant model $D_{H\phi}$ to obtain the final reward, which mainly aims at judging whether there is a hypernym-hyponym relationship between anchor concept and generated query concept. }
%     \label{fig:framework.png}
% \end{figure}

\section{Problem Formalization}
In this section, we first define the preliminaries used in this paper, then formally define the taxonomy entering evaluation task and traditional taxonomy expansion problems.

\subsection{Preliminary and Problem Definition} \label{sec:define}

\begin{myDef}
\textbf{(Taxonomy and Anchor Concept.)}
A taxonomy $\mathcal{T}=(\mathcal{A}, \mathcal{E})$ is a directed acyclic graph.
Each node $a \in \mathcal{A}$ denotes an anchor concept and each directed edge $\langle a_{p}, a_{c} \rangle \in \mathcal{E}$ represents the hyponymy relation between the parent (hypernym) concept $a_p$ and the child (hyponym) concept $a_c$.
\end{myDef}

\begin{myDef}
\textbf{(Query Concept.)}
Query concept is the concept prepared to be added to the existing taxonomy.
The set of query concepts is represented as $\mathcal{Q} = \{q_1, q_2, \cdots, q_m\}$.
\end{myDef}

\begin{myproblem}

\textbf{Taxonomy entering Evaluation}: 
\label{taxonomy entering evaluation}
In taxonomy entering evaluation task, the input is (1) an existing taxonomy $\mathcal{T}^0=(\mathcal{A}^0, \mathcal{E}^0)$, (2) a set of query concepts $\mathcal{Q}$.
The output of entering evaluation is a set of clean query concepts $\mathcal{Q}^*$, which is formalized into the following formula:
\begin{equation}
\small
    \mathcal{Q}^*=\mathcal{Q},\ \forall q\in \mathcal{Q}^* \ s.t.\ P(q|\mathcal{T}^0;\Theta_2)>\gamma.
\end{equation}
$\Theta_2$ denotes the parameters, and $\gamma$ is a given threshold.

\end{myproblem}

\begin{myproblem}
\textbf{Taxonomy Expansion}: 
\label{taxonomy expansion}
In taxonomy expansion task, the input is (1) an existing taxonomy $\mathcal{T}^0=(\mathcal{A}^0, \mathcal{E}^0)$, (2) a set of clean query concepts $\mathcal{Q}^*$.
The output of the taxonomy expansion is a best-expanded taxonomy $\mathcal{T}^*$ with all the clean query concepts added in:
\begin{equation}
\small
\begin{aligned}
\mathcal{T}^{*} &=\underset{\mathcal{T}}{\arg \max } \mathbf{P}(\mathcal{T} \mid \mathcal{T}^0; \Theta) \\
&=\underset{\mathcal{T}}{\arg \max } \sum_{i=1}^{\left|\mathcal{A} \cup  \mathcal{Q}^*\right|} \log \mathbf{P}\left(q_i \mid p\left(q_i\right); \Theta\right),
\end{aligned}
\end{equation}
where $\Theta$ denotes model parameters, and $p(q_i)$ denotes the parent of $q_i$ in the taxonomy $\mathcal{T}$.
\end{myproblem}

\section{Framework of GANTEE}
\label{03}

GANTEE consists of a generator $G$ and two discriminators, which are hyper discriminator $D_H$ and rollout discriminator $D_R$. 
$G$ takes an anchor concept and a label that generates positive or negative samples as inputs and generates the corresponding query concept. 
$D_H$ determines whether the given anchor concept is suitable for the generated query concept based on the given label, and $D_R$ determines whether the generated text is a concept.

Specifically, when a new query concept is given in the predicting stage, GANTEE first determines if it is a concept by $D_R$. 
Then GANTEE determines whether the given concept has a hypernym-hyponym relation to the root concept by $D_H$.
Finally, GANTEE output the confidence score of if the given query concept is suitable for this taxonomy.

\begin{figure}
    \centering
    
    \resizebox{1\linewidth}{!}{\includegraphics{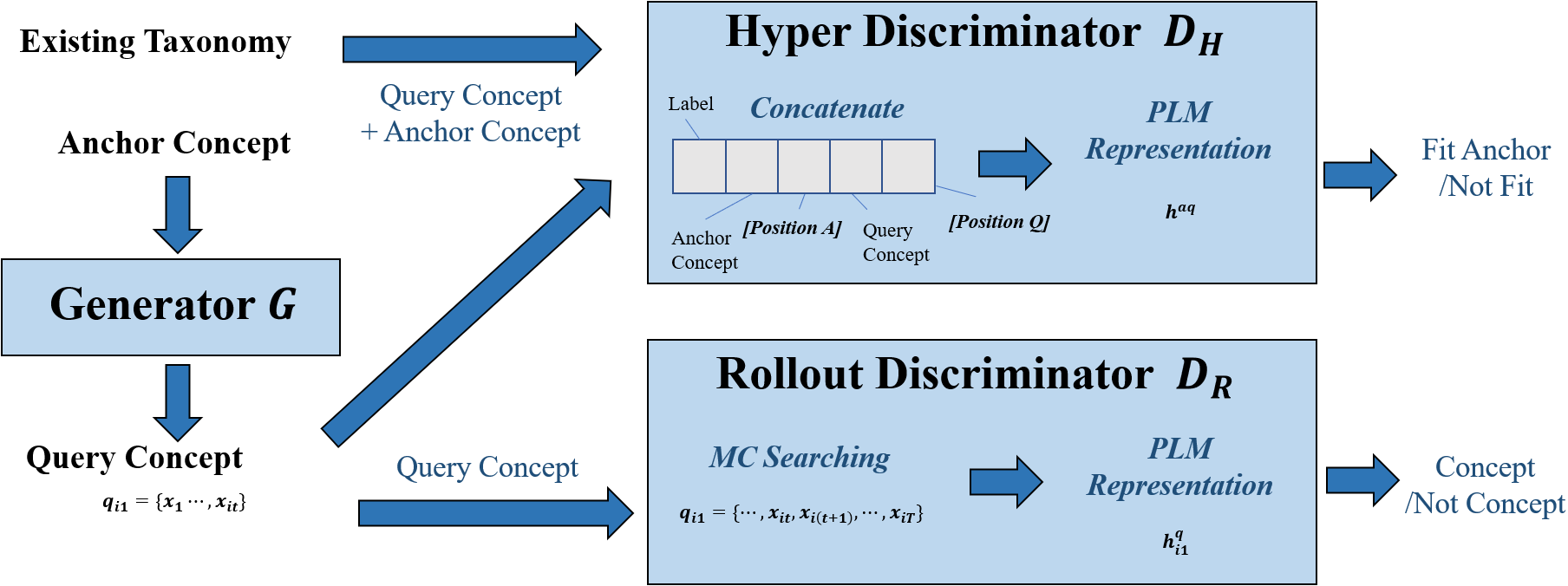}}
    \caption{The framework of GANTEE.
    The generator of the GANTEE is on the left, and duel discriminator is on the right.}
    
    \label{fig:framework.png}
    \vspace{-6mm}
\end{figure}
\subsection{The Generative Model for Query Concept Generating}
We use Transformer~\cite{vaswani2017attention} as the generative model.
Transformer maps the input sequence $x_1,\dots,x_n$ into a sequence of hidden states $h_1,\dots,h_n$.
Specifically, a Transformer block consists of stacked self-attention layers with residual connections.
Each self-attention layer receives $n$ embeddings $\{h^{(l)}_i\}^n_{i=1}$ corresponding to unique input tokens, and outputs $n$ hidden states $\{h^{(l+1)}_i\}^n_{i=1}$.
The $i$-th token is mapped via linear transformations to a key $k_i$, query $q_i$, and value $v_i$.
The $i$-th output of the self-attention layer is given by weighting the values $v_j$ by the normalized dot product between the query $q_i$ and other keys $k_j$:
\begin{equation}
\small
    h_i=\sum^n_{j=1}softmax(\langle q_i,k_{j'}\rangle^n_{j'=1})_j\cdot v_j.
\end{equation}

This allows the layer to assign ``credit'' by implicitly forming state-return associations via similarity of the query and key vectors (maximizing the dot product). 
It is worth noticing that most of the Transformer variants, such as BART~\cite{lewis2019bart} or XLNet~\cite{yang2019xlnet}, and the RNN variants, such as the Gated Recurrent Unit(GRU)~\cite{cho2014gru} or Long-Short Term Memory network(LSTM)~\cite{hochreiter1997lstm}, can be used as a generator in GANTEE.

We use GPT-2~\cite{radford2019gpt2} in this paper to give the best experimental results.
GPT-2 modifies the transformer architecture with a causal self-attention mask to enable auto-regressive generation. 
It is one of the most famous auto-regressive models, which best suits the fake query generation task.

As an auto-regressive generative model, GPT-2 could generate sequences based on all input sequences.
So we formulate the input and the output of the GPT-2 in the following format:
\begin{equation}
    GPT2(l_i\oplus a_i) = l_i\oplus a_i\oplus q_i.
\end{equation}
$GPT2$ denotes the GPT-2 model, $l_i$ denotes the given label, $a_i$ denotes the text of the given anchor concept, $q_i$ denotes the text of the given query concept, and $\oplus$ denotes the concatenation of the tokens.

\subsection{The Rollout Discriminative Model for Short Term Reward in Generating}
The rollout discriminative model is used to judge whether the generated text is a concept before the generating process finish.
Deep discriminative models such as deep neural network(DNN), convolutional neural network(CNN), and recurrent convolutional neural network(RCNN) have shown their effectiveness in complicated sequence classification tasks~\cite{guo2018leakgan, yu2017seqgan, nie2018relgan}.
In this paper, we choose LSTM~\cite{hochreiter1997lstm} as our rollout discriminator.
Since previous study~\cite{guo2018leakgan} find that severe gradient vanishing occurs when the discriminator is too strict, LSTM is a lightweight model which is easy to do the training and will not provide too strict discriminative ability.
LSTM will take in the input embedding $x_1,\dots,x_t$ and use purpose-built memory cells to calculate the hidden state for each step $h_1, \cdots, h_t$.
\begin{equation}
\small
    \begin{aligned}
    \{h_1,\cdots,h_t\} = lstm(x_1,\cdots,x_t),
    \end{aligned}
\end{equation}
where $lstm$ denotes the model of LSTM.
We take the final hidden state $h_t$ and inject it into a linear layer to calculate the reward of the generated text:
\begin{equation}
\small
    r_r = W_r \cdot h_t+b_r,
\end{equation}
where $W_r$ and $b_r$ are the parameters in a linear layer, the $r_r$ is the confidence of whether the generated text is a concept.

\subsection{The Hyper Discriminative Model for Long Term Reward in Generating}
The hyper discriminative model is used to judge whether the generated concept has a hypernym relation to the given anchor concept after the generating process finish based on the given label.
Such discriminative models offer a long-term reward for the generating process to guide the direction of the generation better.
The existing models for taxonomy expansion can be used as the hyper discriminative model, which can detect whether there is a hypernym relation between two concepts.

However, these models rely heavily on the vector representation of the concepts, many studies such as STEAM~\cite{yu2020steam} or HiExpan~\cite{shen2018hiexpan} use external information to get better representations for the concepts, and that would be time-consuming for the generating process since each time the generative model produce a new concept, it will take time to get its detailed external information and get its vector representation.

Instead, we use Bert~\cite{devlin2018bert} to retrieve the representation of concepts since it shows its ability in Dependency Judgement task~\cite{shi2019next}, which is also good at understanding the hypernym (hyponym) relation between two concepts.
We concatenate the label representation and the position embedding proposed by the TaxoExpan~\cite{shen2020taxoexpan} to the Bert representation and put them into a linear layer to get the final predicted result.

Specifically, we get the representation of the concept by Bert in the following format:
\begin{equation}
\small
% \resizebox{0.7\columnwidth}{!}{
\begin{aligned}
    &x_{a}=Bert(x^{c_0}_1,\dots,x^{c_0}_t)[1],\\
    &x_{q}=Bert(x_1,\dots,x_t)[1],\\
    &l = EMB(positive)\text{ or }EMB(negative).
\end{aligned}
% }
\end{equation}
$Bert$ denotes the BERT model, $EMB$ denotes the embedding model, and $positive$ or $negative$ denotes the input to the embedding model.

Then, we concatenate the label representation and position embedding to the representation of the anchor concept and the query concept and put them into a linear layer:
\begin{equation}
\small
    r_h=W_h\cdot(l+x_{a}+p_{a}+x_{q}+p_{q}) + b_h
\end{equation}
Where $W_h$ and $b_h$ are the parameters in a linear layer, $p_{a}$ and $p_q$ are the position embedding. 
The $r_h$ is the probability of the hypernym relation between the generated query concept and the anchor concept.

\subsection{Training for GANTEE}
GANTEE uses a Generative Adversarial Network (GAN) model to combine taxonomy entering evaluation and high-quality training samples generating.
It needs a generative model to produce conceptual text and use discriminative models to offer supervised signals to the generative model.
However, naive GAN is used for continuous data, not discrete tokens.
The previous researchers~\cite{bachman2015data, bahdanau2016actor, yu2017seqgan} propose considering the text generation procedure as a sequential decision-making process.
The objective function of such generative models is formulated to generate a sequence from the start state $s_0$ to maximize its expected end reward via policy gradient algorithms~\cite{sutton1999policy}:
\begin{equation}
    \begin{aligned}
        J(\theta)&=\mathbb{E}[R_T|s_0=l\oplus a;\theta]\\&=\sum_{y_1\in Y}G(y_1|s_0=l\oplus a)\cdot Q^{G}_{D}(s_0=l\oplus a, y_1),
    \end{aligned}
\end{equation}
where $J(\theta)$ is the object function, $R_T$ is the reward for a complete concept that has hypernym relation to the given anchor concept $a$ and label $l$.
Note that the reward is from both the hyper discriminator $D_{H}$ and rollout discriminator $D_{R}$.
$G$ is the generative model, and $Q^{G}_{D}(s, a)$ is the action-value function of a sequence, which is calculated in the following format:
\begin{equation}
    Q^{G}_{D}(a=y_t, s=Y_{1:t-1}, s_0) = D(s_0, Y_{1:t-1})
\end{equation}
Since the reward getting from an unfinished sequence often have a large bias over the finished one, the Monte Carlo search is used to sample the unknown last $T-t$ tokens~\cite{yu2017seqgan}:
\begin{equation}
        \{Y^1_{1:T},\dots,Y^N_{1:T}\}=MC^{G}\left(Y_{1:t};N\right)
\end{equation}

We use rollout discriminator $D_{R}$ to calculate the expected reward based on Monte Carlo search during the generative process for short-term reward and use hyper discriminator $D_{H}$ to calculate the reward when the generating is done for long-term reward. 
The action-value function is transformed into the following format:
\begin{equation}
    \small
    \begin{aligned}
    &Q^{G}_{D}(a=y_t, s=Y_{1:t-1}, s_0) = \\
    &\begin{cases}
    \frac{1}{N}\sum^N_{n=1}D_{R}(s_0, Y_{1:t-1}), Y^n_{1:T}\in MC^{G}\left(Y_{1:t};N\right)& \text{for }t<T\\
    D_{H}(Y_{1:t}, s_0)& \text{for }t=T
    \end{cases}
    \end{aligned}
\end{equation}

Once we have a set of more realistic generated sequences, we will re-train the discriminator model as follows:
\begin{equation}
    \label{eqt:object fucntion}
    \min_{D}-\mathbb{E}_{Y\sim p_{data}}[\log D(Y)]-\mathbb{E}_{Y\sim G}[log(1-D(Y))]
\end{equation}

% \subsection{Implementation Techniques}
Specifically, we pre-train the generator and the discriminator before the adversarial process begins.
After the pre-training stage, the generator and discriminator have trained alternatively.
As the generator is updated via some training epochs, the discriminators need to be re-trained periodically to keep a good pace with the generator.

When training the discriminator, positive samples are sampled from the existing taxonomy $\mathcal{T}^{0}$ based on the hypernym relation $\mathcal{R}$. 
In contrast, negative samples are the fake samples generated from our generator.
To ensure the learning process is doing well, we use both generated samples and pure noisy samples as the negative query concepts to inject into the training process. 
The number of negative query concepts we use is the same as the size of positive query concepts.

\section{Experiments}
\begin{table*}[!t]
    \centering
    \resizebox{0.95\textwidth}{!}{
    \begin{tabular}{c|cccc|ccc|ccc}
        \toprule
        \multirow{2}{*}{\textbf{Method}} & \multicolumn{4}{c|}{\textbf{MAG-CS (10k-level)}} & \multicolumn{3}{c|}{\textbf{MAG-FoS (100k-level)}} & \multicolumn{3}{c}{\textbf{CN-Probase (1000k-level)}} \\
         & MR & MRR@10 & hit@5 & time & MR & MRR@100 & time & MR & MRR@1k & time \\
        \midrule
% 'MR': 2259.935, 'hit@1': 0.0278, 'hit@3': 0.0232, 'hit@5': 0.0124, 'MRR': 0.13688529366800115
% {'MR': 23193.77, 'hit@1': 0.02, 'hit@3': 0.02, 'hit@5': 0.012, 'MRR': 0.12806979592017617}
% {'MR': 2453.602, 'hit@1': 0.002, 'hit@3': 0.0, 'hit@5': 0.0, 'MRR': 0.0022523345576482764}
% MAG_FoS {'MR': 25765.254901960783, 'hit@1': 0.0, 'hit@3': 0.0, 'hit@5': 0.0, 'MRR': 0.0014624484910608182}
% MAG_FoS {'MR': 25645.110638297872, 'hit@1': 0.0, 'hit@3': 0.012, 'hit@5': 0.023, 'MRR': 0.0015337247888280332}
        Close-Position   & 2.3k & 0.137 & 0.023 & - & 23.2k & 0.128 & - & 157.5k & 0.159 & -  \\
        Close-Neighbor & 1.8k & 0.153 & 0.043  & - & 15.6k & 0.146 & -  & 95.5k & 0.188 & -  \\
        BERT+MLP  & 2.5k & 0.001 & 0.002 & 116.0 s & 25.6k & 0.023 & 5.4k s & 159.1k & 0.143 & 42.5k s  \\
        % Octet & 1 & 1 & 1 & 1 & 1 & 1 & 1 & 1 & 1 & 1\\
        % QEN & 1 & 1 & 1 & 1 & 1 & 1 & 1 & 1 & 1 & 1\\
        TaxoExpan  & 1.3k & 0.319 & 0.157 & 471.4 s & 3.7k & 0.305 & 25.2k s & - & - & (200k+) s  \\
        % TMN  & - & - & - & - & - & - & - & - & - & - & - & -  \\
        \midrule
        TaxoExpan & 
        \multirow{2}{*}{1.2k} & \multirow{2}{*}{0.360} & \multirow{2}{*}{0.193} & \multirow{2}{*}{152.8 s} &
        \multirow{2}{*}{2.5k} & \multirow{2}{*}{0.382} & \multirow{2}{*}{6.1k s} &
        \multirow{2}{*}{6.6k} & \multirow{2}{*}{0.319} & \multirow{2}{*}{67.4k s}\\ + GANTEE&&&&&&&&&\\
        
        % Octet & \multirow{2}{*}{1} & \multirow{2}{*}{1} & \multirow{2}{*}{1} & \multirow{2}{*}{1} & \multirow{2}{*}{1} & \multirow{2}{*}{1} & \multirow{2}{*}{1} & \multirow{2}{*}{1} & \multirow{2}{*}{1} & \multirow{2}{*}{1}\\
        %  + GANTEE&&&&&&&&&\\
        % QEN & \multirow{2}{*}{1} & \multirow{2}{*}{1} & \multirow{2}{*}{1} & \multirow{2}{*}{1} & \multirow{2}{*}{1} & \multirow{2}{*}{1} & \multirow{2}{*}{1} & \multirow{2}{*}{1} & \multirow{2}{*}{1} & \multirow{2}{*}{1}\\
        %  + GANTEE&&&&&&&&&\\
        
        \bottomrule
    \end{tabular}
    }
    \caption{Overall Performance on Taxonomy Expansion.}
    \vspace{-5mm}
    \label{tab:overall}
\end{table*}

There is an open question that whether the generated positive samples should be treated as positive training samples or negative training samples.
Since the purpose of GAN is to train a generator that output a distribution similar to the real data, if all the data generated by the generator are regarded as negative samples, it will beyond the discriminative ability of the classifier in downstream applications.
In this paper, we treat them as negative training samples in the first epoch of adversarial training and as positive training samples in the subsequent epochs of adversarial training.

\subsection{Experimental Setup}
\subsubsection{Dataset}
We study the performance of GANTEE on three large-scale real-world taxonomies:
\begin{itemize}
    \item \textbf{Microsoft Academic Graph on Field-of-Study (MAG-FoS):} This taxonomy~\cite{sinha2015mag} consists of public Field-of-Study Taxonomy(FoS).
    It contains over 0.6 million scientific concepts and more than 0.7 million taxonomic relations in English.
    \item \textbf{Microsoft Academic Graph on Computer-Science (MAG-CS):} Following the work of TaxoExpan~\cite{shen2020taxoexpan}, we construct MAG-CS based on the sub-graph of MAG-FoS related to the ``Computer Science'' domain.
    \item \textbf{CN-Probase:} This is an open chinese general concept taxonomy CN-Probase~\cite{chen2019cnprobase}.
    It contains over 300 thousand concepts and more than 30 million ``isA'' relations between concepts and entities in Chinese.
\end{itemize}
Notice that there are many outlier concepts in the taxonomy, so we remove all concepts that do not relate to the original taxonomy.
To ensure the validity and test data won't be trained, we randomly mask 20\% of leaf concepts (along with their relations) for validation and testing.
Furthermore, we add noisy query concepts as 60\% size of leaf concepts into the validation and test set to conduct the experiments as in the real-scenarios. 

\subsubsection{Evaluation Metric}
\begin{itemize}
    \item \textbf{Mean Rank (MR)}: measures the average rank position of a query concept’s true parent among all candidates. 
    % For queries with multiple parents, we first calculate the rank position of each parent and then take the average of all rank positions. 
    % A smaller MR value indicates better model performance.
    \item \textbf{Mean Reciprocal Rank@k (MRR@k)}: calculates the reciprocal rank of a query concept’s true parent. We use a scaled version of MRR in the below equation:
    $$MRR=\frac{1}{|C|}\sum_{c\in C}\frac{1}{|parent(c)|}\sum_{i\in parent(c)}\frac{1}{R_{i,c}/k}$$
    where $parent(c)$ represents the parent node set of the query concept $c$, and $R_{i,c}$ is the rank position of query concept c’s true parent $i$. 
    % We scale the original MRR by a factor from 10 to 1000 (1k) to amplify the performance gap between different methods.
    \item \textbf{hit@k}: is the number of query concepts whose parent is ranked in the top k positions, divided by the total number of queries.
    \item \textbf{time}: denotes the seconds used in the predicting stage of each model on each dataset.
\end{itemize}

\subsubsection{Baseline Methods}
We compared our proposed GANTEE model with the following baseline approaches:
\begin{itemize}
    \item \textbf{Closest-Position}: A rule-based method that chooses the anchor concept which have the most similar embedding with query concept as the query concept’s parent.
    % A rule-based method that first scores each candidate position in the existing taxonomy based on its cosine distance to the query concept between their initial embedding and then ranks all positions using this score. 
    \item \textbf{Closest-Neighbor}: Another rule-based method that scores each position based on its distance to the query concept plus the average distance between its children nodes and the query.
    \item \textbf{BERT+MLP}: This method utilizes BERT~\cite{devlin2018bert} to perform hypernym detection.
    This model's input is the term's surface name, and the representation of BERT's classification token $\langle CLS\rangle$ is fed into a feed-forward layer to score whether the first sequence is the ground-truth parent.
    \item \textbf{Random}: This method is used only in the taxonomy entering evaluation task, which is random to determine if the query concept should be added to the existing taxonomy or not.
    \item \textbf{TaxoExpan}~\cite{shen2020taxoexpan}: One state-of-the-art taxonomy expansion framework which leverages position enhanced graph neural network to capture local information and InfoNCE~\cite{oord2018representation} loss for robust training.
    % \item \textbf{GANTEE}: This is our proposed pluggable framework, which is used to exclude the noisy query concepts and generate more high-quality training data for the taxonomy expansion task.
\end{itemize}
Notably, except for the rule-based method Closest-Position and Closest-Neighbor, other baselines are learning-based methods and designed for one-to-one matching.   
There are other recently proposed taxonomy expansion methods, e.g., HiExpan~\cite{shen2018hiexpan} and STEAM~\cite{yu2020steam}. 
We do not include them as baselines because they leverage external sources, e.g., text corpus, to extract complicated features. 
In contrast, TaxoExpan and other baselines only take initial feature vectors as input.

\subsubsection{Parameter Setting}
For learning-based methods, we use SGD optimizer with initial learning rate 0.0001 and ReduceLROnPlateau\footnote{https://pytorch.org/docs/stable/optim.html/\#torch.optim.lr\\\_scheduler.ReduceLROnPlateau} scheduler with ten patience epochs. 
During model training, the batch size and negative sample size are set to 16 and 256 in the overall performance experiments, respectively. 
We set the epochs to be ten and use two layers of Graph Attention Layer with 8 and 1 attention head and 100 dimensions of the position dim.

\begin{figure*}[h]
	\centering
	\begin{minipage}[h]{0.23\textwidth}
		\centering
		\includegraphics[width=1.1\linewidth]{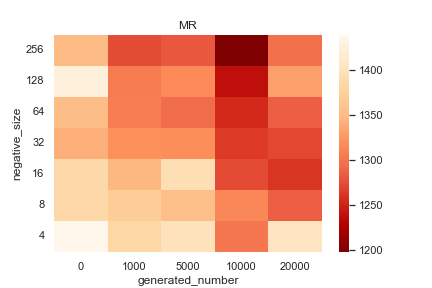}
	\end{minipage}
	\begin{minipage}[h]{0.23\textwidth}
		\centering
		\includegraphics[width=1.1\linewidth]{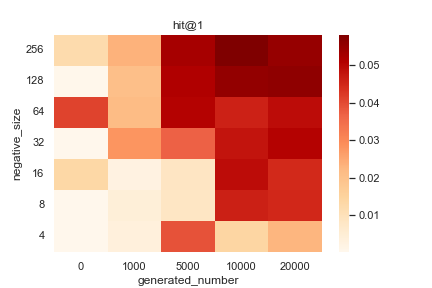}
	\end{minipage}
	\begin{minipage}[h]{0.23\textwidth}
		\centering
		\includegraphics[width=1.1\linewidth]{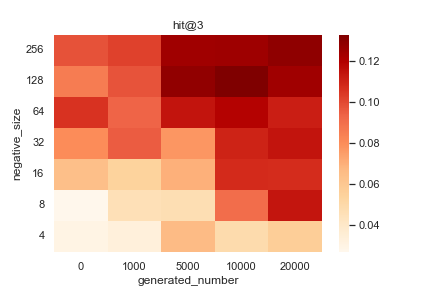}
	\end{minipage}
	\begin{minipage}[h]{0.23\textwidth}
		\centering
		\includegraphics[width=1.1\linewidth]{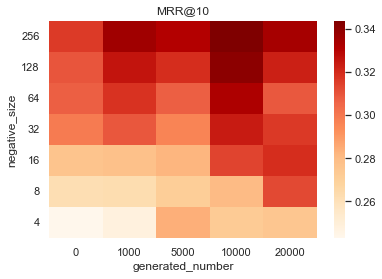}
	\end{minipage}
	\caption{The changing of the results on different metrics with different negative size sampling during the taxonomy expansion model training stage and different number of generated high-quality samples by GANTEE.}
	\label{Fig.detail} 
	\vspace{-3mm}
\end{figure*}

\begin{table}[!t]
    \centering
    \resizebox{0.9\columnwidth}{!}{
    \begin{tabular}{c|cc|cc}
        \toprule
        \multirow{2}{*}{\textbf{Method}} & \multicolumn{2}{c|}{\textbf{MAG-CS}} & \multicolumn{2}{c}{\textbf{CN-Probase-Sampled}} \\
         & Acc & F1 & Acc & F1 \\
        \midrule
        Closest-Position  & 21.5 & 13.9 & 15.4 & 11.5 \\
        Closest-Neighbor  & 28.8 & 17.8 & 19.6 & 13.7 \\
        BERT+MLP  & 57.6 & 46.1 & 52.3 & 43.8 \\
        Random & 74.8 & - & 75.9 & - \\
        TaxoExpan  & 76.4 & 52.7 & 66.8 & 32.1 \\
        % TMN  & - & - & - & - \\
        \midrule
        GANTEE  & \textbf{89.2} & \textbf{79.3} & \textbf{77.5} & \textbf{65.4}\\
        \bottomrule
    \end{tabular}
    }
    \caption{Performance on Taxonomy Entering Evaluation.}
    \label{tab:tee}
\end{table}

\begin{figure*}[h]
	\centering
	\begin{minipage}[h]{0.23\textwidth}
		\centering
		\includegraphics[width=1.05\linewidth]{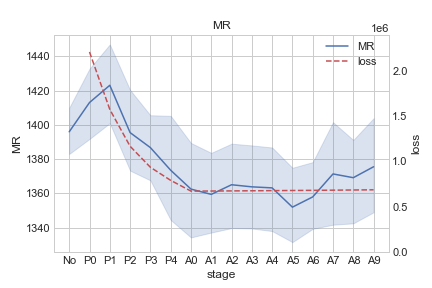}
	\end{minipage}
	\begin{minipage}[h]{0.23\textwidth}
		\centering
		\includegraphics[width=1.05\linewidth]{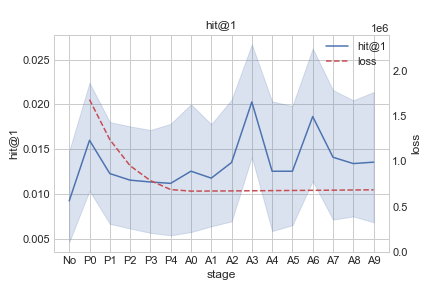}
	\end{minipage}
	\begin{minipage}[h]{0.23\textwidth}
		\centering
		\includegraphics[width=1.05\linewidth]{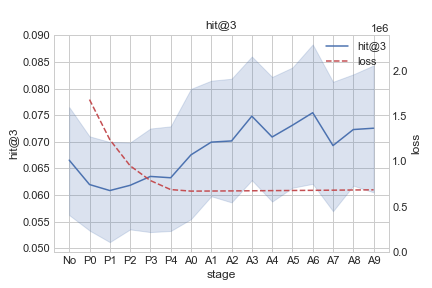}
	\end{minipage}
	\begin{minipage}[h]{0.23\textwidth}
		\centering
		\includegraphics[width=1.05\linewidth]{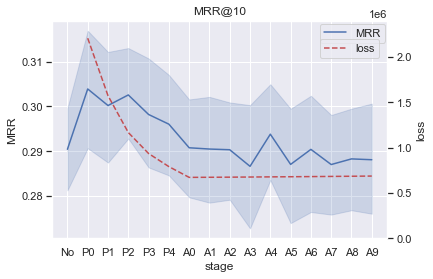}
	\end{minipage}
	\caption{The changing of the losses and results on different metrics with the adversarial training goes on. In these figures, red lines denote ``loss'', and blue line denotes the metrics, label ``No'' means taxonomy expansion without GANTEE, label ``P'' means the stage of pretraining generative model, and label ``A'' means the adversarial training stage.}
	\label{Fig.stage} 
    \vspace{-5mm}
\end{figure*}

\begin{table}[!t]
    \centering
    
    \resizebox{0.9\columnwidth}{!}{
    \begin{tabular}{c|ccc|cc}
        \toprule
        \textbf{Taxonomy} & \multicolumn{3}{c|}{\textbf{MAG-CS TE}} & \multicolumn{2}{c}{\textbf{MAG-CS TEE}}\\
        \textbf{+ GANTEE} & MR & MRR & hit@5 & Acc & F1 \\
        \midrule
        w/o hd  & 1235 & 0.357 & 0.186 & 78.5 & 62.9\\
        w/o rd  & 1217 & 0.351 & 0.182 & 81.5 & 71.3\\
        w/o $D_\phi$  & 1250.1 & 0.348 & 0.180 & 71.4 & 51.9\\
        - & \textbf{1188.1} & \textbf{0.360}& \textbf{0.193} & \textbf{89.2} & \textbf{79.3} \\
        \bottomrule
    \end{tabular}
    }
    \caption{Ablation study on GANTEE.}
    \label{tab:ablation}
    \vspace{-6mm}
\end{table}
\subsection{Experimental Results}

\subsubsection{Overall Performance on Taxonomy Expansion}
As our proposal is a plugin framework for improving both the effectiveness and efficiency of the taxonomy expansion, we are curious how GANTEE will affect the performance of the previous methods on the taxonomy expansion task. 
Thus, we use GANTEE on the most prevailing State-Of-The-Art (SOTA) method TaxoExpan~\cite{shen2020taxoexpan} on the taxonomy expansion task. 
The results are presented in Table~\ref{tab:overall}.

Notably, since the time expense of the rule-based method only depends on the computational complexity of the algorithms or matrix operations, it is not recorded, for it can be neglected most of the time in our experiments. 
And the time expense of TaxoExpan on CN-Probase is too large, and we do not finish this experiment in days, so the expected time expense is recorded in the table.

\textbf{Effectiveness Performance:}
From the results, we can see that all of the rule-based methods did a poor job in most metrics, which demonstrates that the semantics of concepts in real-scenarios are complicated, so simple rule-based methods have difficulty performing well in the taxonomy expansion task.
However, BERT+MLP has also done a poor job on most metrics, demonstrating that simply using the representation of PLM is not suitable for taxonomy expansion tasks.
With our GANTEE, TaxoExpan can reach a new SOTA in all three datasets.
Additionally, the larger scale of the taxonomy, the more improvement is made through GANTEE.  

\textbf{Efficiency Performance:}
BERT+MLP method is proven itself to be an efficient method, and we think it is because this method also uses PLMs to fetch the representation of each concept, which is a trade to the final effectiveness of the taxonomy expansion task.
Moreover, when TaxoExpan is implemented after our proposed method, a new SOTA is reached within one-third of the predicting time on three datasets on average.

\subsubsection{Performance on Taxonomy Entering Evaluation}
Table~\ref{tab:tee} presents the results of all compared methods on the three datasets. 
The rule-based methods and BERT+MLP did a poor job on taxonomy entering evaluation task, which is much worse than the Random. 
The TaxoExpan has limited ability to distinguish the noisy query concept, for it outperforms four of the other baseline methods.
GANTEE outperforms all the other methods by a large margin on all datasets in taxonomy entering evaluation task.
Furthermore, since the PLM representation of the English word is better than the Chinese word, it is easier for the discriminator to distinguish the noisy query concepts in English than in Chinese, which makes all the metrics in English higher than in Chinese.

\subsubsection{Ablation Study}

The detail of ablation studies is presented in this subsection. 
We remove three of the essential designs in the GANTEE. 
In ``w/o hd'', we remove the hyper discriminator during the adversarial generating process. 
In ``w/o rd'', we remove the rollout discriminator during the adversarial generating process. 
In ``w/o $D_\phi$'', we remove all the adversarial processes and only use the generator to generate samples. 
The results are shown in Table~\ref{tab:ablation}

We compare the removal of the hyper discriminative model $D_H$ and rollout discriminative model $D_R$ together.
We can see that without the rollout discriminative model, all metrics have a more significant decline than without the hyper discriminative model in the taxonomy expansion task.
However, when the hyper discriminative model is removed, there will be a more significant decline in the metrics in the taxonomy entering evaluation task than removing the rollout discriminative.
We conclude that the rollout discriminative model generates better conceptual texts, and the hyper discriminative is used to strictly denoise all the noisy query concepts.

And without all of the discriminative models, which means only using the generative model to learn from the query conceptual text, a significant decline is witnessed in all metrics. 
We conclude there are two main reasons.
The first reason for this significant decline is because there has no model trained for taxonomy entering evaluation specifically, so the metrics of taxonomy entering evaluation task is decreasing.
The second reason for this significant decline is because there have no supervised signals provided during the generating process, and the GPT-2 is hard to capture the semantic of the hypernym-hyponym relationship in the existing taxonomy, so the result of the taxonomy expansion task is decreasing.

\subsubsection{Analysis of GANTEE}

The Table~\ref{Fig.detail} denotes how different the negative size during training and a different number of the high-quality generated training sampled by GANTEE affect the final result of the taxonomy expansion task.
The upper right corner of the figure is significantly darker than the other areas. 
So we conclude that when generating more high-quality training samples by GANTEE, the taxonomy expansion task will improve performance by using a bigger negative size of the taxonomy expansion in its training stage.

The Table~\ref{Fig.stage} denotes how the final result of the taxonomy expansion task will be affected when the GANTEE stop training in a different training stage.
In these figures, we can intuitively conclude that with the stage of the training going deeper, the performance of the final result will not be better because the line in MRR, hit@1, and hit@3 are not consistently decreasing, and the line in MR is not continuously increasing.
We conclude that when the loss no longer drops dramatically, it is better to stop the training of the GANTEE to receive a better performance in the taxonomy expansion task.
We find out that the goal of the generative model in GANTEE is to confuse the discriminator, and when the distribution of the generated text is close to the distribution of the real conceptual text, the discriminator will not be able to identify the samples effectively. 
And the gradients are not effectively obtained, causing the discriminator to become less and less effective, affecting the quality of the generator and generated training data.

% \begin{table}[!t]
%     \centering
%     \begin{tabular}{c|cccc}
%         \toprule
%         \multirow{3}{*}{Error 1}   
%         & \multicolumn{4}{c}{\textbf{Label:} Positive }\\  
%         & \multicolumn{4}{c}{\textbf{Given Anchor:} mackerel sharks }\\ 
%         & \multicolumn{4}{c}{\textbf{Generated:} mackerel sharks }\\ 
%         Error 3 & N & 46.1 & 52.3 & 43.8 \\
%         \bottomrule
%     \end{tabular}
%     \caption{Performance on Taxonomy Entering Evaluation.}
%     \label{tab:tee}
%     \vspace{-2mm}
% \end{table}

\begin{table}[!t]
    \centering
    \small
    \begin{tabular}{|l|}
        \makecell[|c|]{}\\
        \hline
        \textbf{Samples Generated by GANTEE} \\
        \hline
        \textbf{Given Label} \\
        \textcolor{purple}{Positive} \\
        \textbf{Given Anchor Concept} \\
        \textcolor{brown}{Deep Learning Algorithm}	 \\
        \textbf{Generated Text} \\
        \textcolor{blue}{Deep Q-Learning} \cmark\\
        \hdashline
        \textbf{Given Label} \\
        \textcolor{purple}{Positive} \\
        \textbf{Given Anchor Concept} \\
        \textcolor{brown}{Computer Vision}	 \\
        \textbf{Generated Text} \\
        \textcolor{blue}{Computer Vision Algorithm} \cmark\\
        \hdashline
        \textbf{Given Label} \\
        \textcolor{purple}{Negative} \\
        \textbf{Given Anchor Concept} \\
        \textcolor{brown}{Deep Learning Algorithm}	 \\
        \textbf{Generated Text} \\
        \textcolor{blue}{Deep Graphics Processing Unit} \cmark\\
        \midrule
        \midrule
        \textbf{Given Label} \\
        \textcolor{purple}{Positive} \\
        \textbf{Given Anchor Concept} \\
        \textcolor{brown}{Deep Q-Learning}	 \\
        \textbf{Generated Text} \\
        \textcolor{red}{Deep Algorithm} \xmark\\
        \hdashline
        \textbf{Given Label} \\
        \textcolor{purple}{Positive} \\
        \textbf{Given Anchor Concept} \\
        \textcolor{brown}{Deep Learning}	 \\
        \textbf{Generated Text} \\
        \textcolor{red}{Deep Graphics Processing Unit} \xmark\\
    \hline
    \end{tabular}
    \caption{Case study of the samples generated by GANTEE. 
    Text in purple and brown represents given label and given anchor concept.
    Text in blue and red denotes good cases and bad cases.}
    \label{tab:case}
    \vspace{-7mm}
\end{table}

\subsubsection{Case Analysis}
We present three good cases and two typical bad cases of GANTEE in Tabel \ref{tab:case}

The result shows that GANTEE has the ability in generating a new concept in a domain-specific field.
For example, ``Deep Q-Learning'', ``Computer Vision Algorithm'' and ``Deep Graphics Processing Unit'' are all correct concepts and have true hypernym-hyponym relationships to the given anchor concepts based on the given label. 

However, we find GANTEE has limited ability in generating positive query concepts for too fine-grained or too coarse-grained anchor concepts.
We conclude that GANTEE has difficulty in capturing the semantics of too fine-grained anchor concepts which are seldom seen in the training process of GANTEE ``Deep Q-Learning Algorithm'' to ``Deep Algorithm''.
And GANTEE tends to assign any concept to the too coarse-grained anchor concepts like ``Deep Graphics Processing Unit'' to ``Deep Learning''.

\section{Related Work}

\subsection{Taxonomy Construction and Expansion}
The previous methods for taxonomy expansion task can be divided into two categories.
The studies in the first category tend to construct a better model to predict the hypernym-hyponym relationships of the concepts.
Position embedding~\cite{shen2020taxoexpan}, concept sorting mechanism~\cite{song2021should} and multilevel position matching~\cite{wang2022qen} are proposed to better understand the semantics of the query concepts and anchor concepts.
The second category focus on learning good representations for all concepts.
Open source~\cite{yu2020steam}, user-clicked-logs~\cite{cheng2022learning} and parent-child information~\cite{wang2021enquire, wang2022qen} are all used by the previous methods to learn a better representation for concepts.
Besides, there exist many studies that use PLMs during the representation learning stage~\cite{cheng2022learning, wang2022qen, takeoka2021low}, however, these methods only use the PLMs as a external information provider to learn a better representation.

\section{Conclusion}
% In this paper, we propose a new task called taxonomy entering evaluation, we mainly used GAN to generate vast negative but confusing samples which should not be added into the existing taxonomy.

% In this paper, we propose an adaptively self-supervised user behavior-oriented product taxonomy expansion framework.
% The user behavior information allows our model to learn the domain-specific relational and structural representations while matching the users' intention and cognition.
% The adaptively self-supervised generation strategy can construct a high-quality and balanced training dataset that avoids inheriting problems in the existing taxonomy.
% Comprehensive experiments conducted on three real-world product taxonomies have shown that the results achieved by our proposed framework improve considerably over state-of-the-art both in automatic and manual evaluations.
% We observe that much other information can be incorporated into the model, such as image and merchant information.
% In the future, we will further explore other means to determine the problematic cases which need commonsense knowledge or domain knowledge.

This paper studies the drawbacks of the existing taxonomy expansion methods when applied to the real-scenarios.
We propose a pluggable framework called GANTEE to alleviate these drawbacks.
Extensive experiments have been conducted to demonstrate that GANTEE can boost the efficiency and effectiveness of the taxonomy expansion task.
Interesting future work includes lower the representation learning expense or learning representation for multi-model queries in the task of taxonomy expansion.

\section{ Acknowledgments}
This work was supported by 
 Shanghai Municipal Special Fund for Promoting High-quality Development of Industries (No. 2021-GZL-RGZN-01018), Shanghai Sailing Program (No. 23YF1409400), % 刘井平
National Key Research and Development Project (No.2020AAA0109302), % 肖仰华
Scientific and technological innovation 2030 - major project of new generation artificial intelligence (No.2020AAA0109300), % 冯红伟
National Natural Science Foundation of China (No.62072323), Shanghai Science and Technology Innovation Action Plan (No. 22511104700). % 李直旭
National Natural Science Foundation of China (No. 62102095). % 梁家卿

% \clearpage
\bibliography{aaai23}

\end{document}